\documentclass[10pt,twocolumn,letterpaper]{article}

\usepackage[final]{cvpr} 
\usepackage{booktabs}
\usepackage{multirow}
\usepackage{array}
\usepackage{graphicx}
\usepackage{makecell}
\usepackage{algorithm}
\usepackage{algpseudocode}
\usepackage{listings}
\usepackage{xcolor}

\usepackage{multirow}
\usepackage[table]{xcolor}
\usepackage{amssymb} 

\definecolor{tableblue}{HTML}{E8F1F7}

\definecolor{cvprblue}{rgb}{0.21,0.49,0.74}
\usepackage[pagebackref,breaklinks,colorlinks,allcolors=cvprblue]{hyperref}

\def\paperID{*****} 
\def\confName{CVPR}
\def\confYear{2026}

\title{Non-Contrastive Vision-Language Learning \\ with Predictive Embedding Alignment}
\author{
Lukas Kuhn$^{*1,2,3}$ \quad
Giuseppe Serra$^{*1,3}$ \quad
Florian Buettner$^{1,2,3}$ \\
{\tt\small \{lukas.kuhn, giuseppe.serra, florian.buettner\}@dkfz-heidelberg.de} \\
\\
$^{1}$Goethe University Frankfurt, Frankfurt, Germany \quad \\
$^{2}$German Cancer Research Center (DKFZ), Heidelberg, Germany \quad \\
$^{3}$German Cancer Consortium (DKTK), Frankfurt, Germany 
}

\begin{document}
\maketitle
\begin{abstract}
Vision-language models have transformed multimodal representation learning, yet dominant contrastive approaches like CLIP require large batch sizes, careful negative sampling, and extensive hyperparameter tuning. We introduce NOVA, a NOn-contrastive Vision-language Alignment framework based on joint embedding prediction with distributional regularization. NOVA aligns visual representations to a frozen, domain-specific text encoder by predicting text embeddings from augmented image views, while enforcing an isotropic Gaussian structure via Sketched Isotropic Gaussian Regularization (SIGReg). This eliminates the need for negative sampling, momentum encoders, or stop-gradients, reducing the training objective to a single hyperparameter. We evaluate NOVA on zero-shot chest X-ray classification using ClinicalBERT as the text encoder and Vision Transformers trained from scratch on MIMIC-CXR. On zero-shot classification across three benchmark datasets, NOVA outperforms multiple standard baselines while exhibiting substantially more consistent training runs. Our results demonstrate that non-contrastive vision-language pretraining offers a simpler, more stable, and more effective alternative to contrastive methods.
\end{abstract}    
\def\thefootnote{*}\footnotetext{These authors contributed equally to this work}\def\thefootnote{\arabic{footnote}}
\section{Introduction}
\label{sec:intro}

\begin{figure}[ht]
  \centering
  \includegraphics[width=0.90\linewidth]{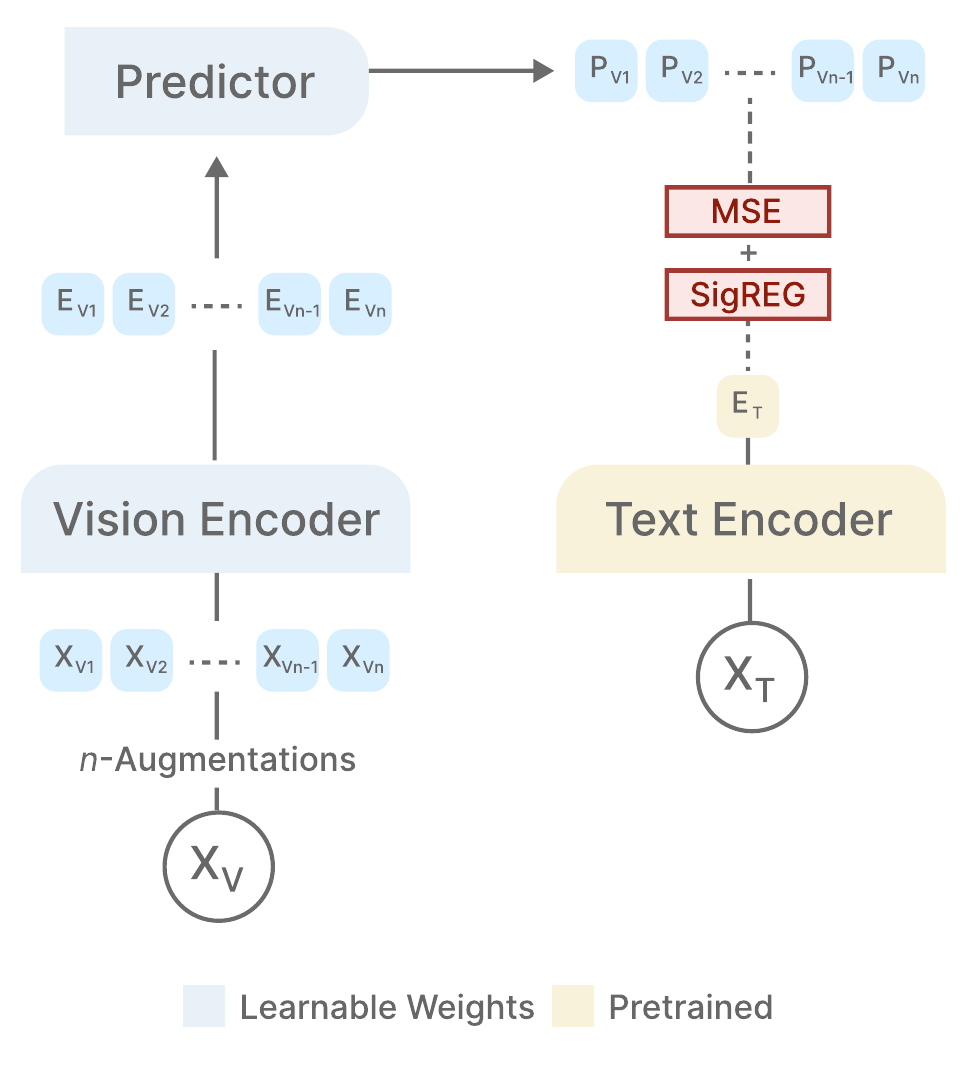}
   \caption{\textbf{NOVA overview.} Given an image-text pair, we generate $n$ augmented views of the image via global and local crops. Each view is encoded by a learnable vision encoder and passed through a learnable predictor to obtain predicted embeddings $\{P_1, \ldots, P_n\}$. The corresponding text is encoded by a frozen, pretrained ClinicalBERT model with a learnable projection head, yielding the target embedding $E_T$. All predicted image views are aligned to the text anchor via MSE loss, while SIGReg regularization enforces an isotropic Gaussian distribution over the joint embedding space to prevent representational collapse. This simple objective enables stable training without negative sampling, momentum encoders, or stop-gradients.}
   \vspace{-1em}
   \label{fig:overview}
\end{figure}

Contrastive learning has emerged as the dominant paradigm for vision-language alignment. CLIP \cite{radford2021clip} demonstrated that training dual encoders with an InfoNCE objective over large-scale image-text pairs yields powerful representations for zero-shot classification, retrieval, and transfer learning. Subsequent work has refined this approach through alternative loss functions \cite{zhai2023siglip}, domain-specific adaptation \cite{wang2022medclipcontrastivelearningunpaired}, and architectural improvements. However, despite these advances, contrastive methods still require careful engineering to succeed: large batch sizes to provide sufficient negative samples, momentum encoders to stabilize training, or asymmetric architectures to prevent collapse. These heuristics add complexity and computational overhead, yet their theoretical justification often remains unclear. Moreover, contrastive learning is best suited to data-rich settings; in low-data regimes typical of medical imaging, for example, the reliance on diverse negative samples becomes a fundamental limitation.

Joint Embedding Predictive Architectures (JEPAs) \cite{lecun2022path} offer an alternative that sidesteps the need for explicit negative sampling. Rather than contrasting positive and negative pairs, JEPAs learn by predicting the embedding of one view from another. LeJEPA \cite{balestriero2025lejepa} provides a theoretically grounded instantiation of this framework, proving that embeddings following an isotropic Gaussian distribution minimize expected downstream risk without requiring stop-gradients, teacher-student networks, or careful hyperparameter scheduling.

In this work, we extend the LeJEPA framework to vision-language alignment. We propose \textbf{NOVA}, which enables the alignment of multiple image representations to text representations encoded by a pretrained text model through embedding prediction with SIGReg regularization. By anchoring visual representations to a fixed linguistic embedding space, we leverage the semantic structure already captured by domain-specific language models while learning visual encoders from scratch. Similar to the original LeJepa, our approach requires only a single hyperparameter controlling the trade-off between alignment and regularization, offering a simple and principled alternative to contrastive vision-language pretraining.
\\
Our contributions are as follows:

\begin{itemize}
    \item We introduce NOVA, a non-contrastive framework for vision-language alignment based on joint embedding prediction with distributional regularization.
    \item We demonstrate that anchoring visual representations to a frozen domain-specific text encoder provides an effective training signal without requiring contrastive sampling or large batch sizes.
    \item We achieve zero-shot classification performance that outperforms both CLIP and MedCLIP baselines across all three benchmarks, using a simpler training objective with highly predictable and consistent training runs.
\end{itemize}

\section{Related work}

\paragraph{Contrastive Vision-Language Learning}
Contrastive Language-Image Pre-Training (CLIP)~\cite{radford2021clip} established the dominant paradigm for aligning visual and textual representations by training dual encoders with a contrastive objective over large-scale image-text pairs. Given a batch of $N$ image-text pairs, let $\mathbf{v}_i$ and $\mathbf{t}_i$ denote the $\ell_2$-normalized embeddings of the $i$-th image and text, respectively. CLIP maximizes agreement between matched pairs while pushing apart unmatched pairs using the InfoNCE loss~\cite{oord2019infonce}:
\begin{equation}
    \mathcal{L}_{\text{InfoNCE}} = -\frac{1}{N} \sum_{i=1}^{N} \log \frac{\exp(\mathbf{v}_i^\top \mathbf{t}_i / \tau)}{\sum_{j=1}^{N} \exp(\mathbf{v}_i^\top \mathbf{t}_j / \tau)}
\end{equation}
where $\tau$ is a learnable temperature parameter. The loss is computed symmetrically for both image-to-text and text-to-image directions. Intuitively, InfoNCE treats the $N-1$ non-matching texts in the batch as negative samples, requiring the model to distinguish the correct pairing from all alternatives. This formulation necessitates large batch sizes to provide a diverse set of negatives, leading to significant computational and memory requirements. 

SigLIP~\cite{zhai2023siglip} replaces the softmax-based contrastive loss with a sigmoid loss operating on image-text pairs independently:
\begin{equation}
    \mathcal{L}_{\text{SigLIP}} = -\frac{1}{N^2} \sum_{i=1}^{N} \sum_{j=1}^{N} \log \sigma\left(y_{ij} \cdot \mathbf{v}_i^\top \mathbf{t}_j \cdot \tau + b\right)
\end{equation}
where $y_{ij} = 1$ if $i = j$ (matched pair) and $y_{ij} = -1$ otherwise, $\tau$ is a learnable temperature, and $b$ is a learnable bias. Unlike InfoNCE, SigLIP treats each pair as an independent binary classification problem, removing the need for global normalization across the batch. This improves scalability and has been shown to yield better performance, particularly at smaller batch sizes. However, both approaches still rely on explicit negative sampling and scale quadratically with batch size in the number of pairwise comparisons.

\paragraph{Self-Supervised Learning in Medical Imaging}
The scarcity of labeled medical imaging data poses a fundamental challenge for supervised deep learning, as expert annotation is time-consuming, costly, and requires specialized clinical knowledge \cite{Shurrab_2022, huang2023self}. Self-supervised learning (SSL) addresses this by learning representations from unlabeled data through pretext tasks, enabling effective transfer to downstream tasks with limited labels. Contrastive methods such as SimCLR \cite{chen2020simple}, MoCo \cite{he2020momentum} and BYOL \cite{grill2020bootstrap} have been widely adopted for medical image analysis, with studies demonstrating improved performance on chest X-ray classification, CT segmentation, and pathology tasks compared to training from scratch or ImageNet initialization \cite{sowrirajan2021mococxrmocopretrainingimproves, azizi2021bigselfsupervisedmodelsadvance}. A systematic review of 79 studies found that contrastive learning was the dominant SSL strategy in medical imaging, with SimCLR, MoCo, and BYOL being the most frequently used frameworks \cite{huang2023self}. Notably, SSL pretraining yields the largest gains when unlabeled examples substantially outnumber labeled ones, a common scenario in clinical settings. Nonetheless, most SSL methods in medical imaging still rely on contrastive objectives that require large batch sizes or momentum encoders to prevent collapse. 

\paragraph{Domain-Specific Language Models}
The success of BERT \cite{devlin2019bertpretrainingdeepbidirectional} on general NLP tasks motivated adaptations to specialized domains. Standard BERT, pretrained on Wikipedia and BookCorpus, performs poorly on clinical text due to the prevalence of medical jargon, abbreviations, and non-standard grammar \cite{huang2020clinicalbertmodelingclinicalnotes}. ClinicalBERT \cite{huang2020clinicalbertmodelingclinicalnotes} addresses this by pretraining on clinical notes from MIMIC-III, achieving substantially higher accuracy on clinical language modeling tasks and better correlation with physician-assessed semantic similarity between medical concepts compared to Word2Vec and FastText. Similarly, BioBERT \cite{lee2019biobert} pretrains on biomedical literature from PubMed, while PubMedBERT \cite{gu2021pubmedpert} demonstrates that pretraining from scratch on in-domain text outperforms continued pretraining of general-domain models. These findings motivate our choice of ClinicalBERT as the frozen text encoder in NOVA: by anchoring visual representations to a text embedding space that already captures clinical semantics, we leverage domain knowledge without requiring the vision encoder to learn it from scratch.

\paragraph{Vision-Language Models in the Medical Domain}
Adapting vision-language models to medical imaging presents unique challenges due to limited paired data and domain-specific terminology. To address data scarcity, MedCLIP~\cite{wang2022medclipcontrastivelearningunpaired} decouples image-text pairs and leverages semantic similarity between text samples, enabling training with unpaired data. Similarly, PubMedCLIP~\cite{eslami2023pubmedclip} pretrains on medical literature figures, while other approaches exploit radiology reports for domain-specific adaptation \cite{zhang2023biomedclip}. While these methods successfully adapt the contrastive framework of CLIP to the medical domain, they also inherit both its strengths and limitations.

\paragraph{Joint Embedding Predictive Architectures.}
Joint Embedding Predictive Architectures (JEPAs)~\cite{lecun2022path} offer an alternative to contrastive learning by predicting the embeddings of one view from another, rather than contrasting positive and negative pairs. Building on this framework, \citet{balestriero2025lejepa} propose LeJEPA, a theoretically grounded variant that proves embeddings following an isotropic Gaussian distribution minimize downstream prediction risk. To enforce this distribution, LeJEPA introduces Sketched Isotropic Gaussian Regularization (SigREG), a computationally efficient alternative to covariance-based regularizers such as VICReg~\cite{bardes2021vicreg}. More recently, VL-JEPA~\cite{chen2025vljepajointembeddingpredictive} extends the JEPA framework to vision-language tasks by predicting continuous text embeddings rather than generating tokens autoregressively. However, this approach still  relies on  a contrastive learning approach and employs the InfoNCE loss with a large-scale frozen vision encoder and learnable pretrained text encoder. VL-JEPA targets video understanding and VQA tasks and relies on massive pretrained encoders (V-JEPA 2 with 304M parameters) and billions of training samples that allow for efficient use of contrastive learning---assumptions that are incompatible with low-data medical imaging settings where training vision encoders from scratch on limited domain-specific data is often necessary.


\paragraph{}
Ultimately, while contrastive multimodal methods have achieved remarkable success, they all share fundamental limitations that become particularly problematic in settings with limited data and resources. In fact, the reliance on explicit negative sampling and the extensive hyperparameter tuning require significant computational overhead to provide sufficient contrast and obtain reasonable results. Thus, motivated by these limitations, we develop \textbf{NOVA} – a \textbf{NO}n-contrastive \textbf{V}ision-language \textbf{A}lignment method which leverages a frozen text encoder to guide SIG-based vision representations into a domain-specific embedding space without prior knowledge. Being non-contrastive, NOVA circumvents these constraints, offering stable training with minimal computational requirements and a single tunable hyperparameter.
\section{Method}

We propose \textbf{NOVA}, a vision-language model built on the LeJEPA framework that aligns visual representations to language through embedding prediction with principled regularization. Unlike classical contrastive approaches that rely on sample-wise comparisons, NOVA directly regresses image embeddings toward a text anchor while enforcing distributional constraints that provably minimize downstream risk.

\subsection{Architecture}

NOVA comprises three components (Figure~\ref{fig:overview}):

\begin{itemize}
    \item \textbf{Vision Encoder} ($X_V \mapsto E_V$): A randomly initialized Vision Transformer encoding augmented image views into visual embeddings.
    \item \textbf{Predictor} ($E_V \mapsto P$): A learnable predictor network that encodes views into predicted embeddings in a shared latent space.
    \item \textbf{Text Encoder} ($Y \mapsto E_T$): A pretrained, frozen text encoder that embeds the corresponding free-text description, followed by a learnable projection head that maps to the predictor's output dimension.
\end{itemize}

Given an image-text pair, we generate $n$ augmented views of the image, producing encoded representations $\{E_{V_1}, \ldots, E_{V_n}\}$ and corresponding predictions $\{P_1, \ldots, P_n\}$. The text is encoded once to yield the target embedding $E_T$; while integrating multiple augmented texts would be straightforward, we leave this extension to future work.

\subsection{Training Objective}

JEPA models optimize two objectives jointly: (1) a prediction error in the embedding space for the alignment of image views and text, and (2) a regularization term to prevent representation collapse \cite{bardes2021vicreg, balestriero2025lejepa} (see Alg. \ref{alg:vllejepa} for a pseudocode implementation). 

\paragraph{Alignment} We align all predicted image views to the text embedding via mean squared error (MSE):
\begin{equation}
    \mathcal{L}_{\text{MSE}} = \frac{1}{n} \sum_{i=1}^{n} \| P_{V_i} - E_T \|_2^2
\end{equation}
By regressing image embeddings toward the frozen text encoder, NOVA leverages the text encoder's established rich semantic structure rather than learning it from scratch. This asymmetry, a frozen text encoder guiding a learnable vision encoder, provides stable optimization targets throughout training, avoiding the co-adaptation dynamics that can destabilize training where both encoders evolve simultaneously.

\paragraph{Regularization} We regularize the joint distribution of all embeddings (i.e., both image predictions and text) using SIGReg \cite{balestriero2025lejepa}. SIGReg enforces an isotropic Gaussian structure by projecting embeddings along random directions and matching the resulting univariate densities via characteristic function tests. This approach prevents both complete collapse (i.e., all embeddings mapping to a single point) and dimensional collapse (embeddings occupying a low-dimensional subspace), without requiring heuristics such as stop-gradients or momentum-based teacher networks. Crucially, SIGReg exhibits linear time and memory complexity, making it scalable to high-dimensional embeddings.

\paragraph{Combined Objective} Combining these two objectives, the full training loss is:
\begin{equation}
    \mathcal{L}_{\text{NOVA}} = (1 - \lambda) \cdot \mathcal{L}_{\text{MSE}} + \lambda \cdot \mathcal{L}_{\text{SIGReg}}
\end{equation}
The hyperparameter $\lambda \in [0, 1]$ balances predictive alignment against distributional regularization. This single hyperparameter replaces the numerous scheduling and balancing terms typically required in self-supervised frameworks, offering a lean and theoretically grounded training objective. This allows the model to learn a joint energy landscape between ($E_T, P_{V_i}$). 

\paragraph{Connection to Energy-Based Models}
NOVA can be interpreted through the lens of energy-based models (EBMs)~\cite{lecun2006tutorial}. The MSE objective defines an energy function $E(V, T) = \| P_V - E_T \|_2^2$ over the joint space of visual and textual representations, where compatible image-text pairs are assigned low energy and incompatible pairs high energy. The text embedding $E_T$ acts as an attractor in the energy landscape, pulling the predicted image views $P_{V_i}$ toward a shared semantic basin. Unlike contrastive methods, which explicitly push apart negative pairs to create energy barriers, NOVA shapes the energy landscape implicitly through SIGReg's distributional constraint, ensuring that the learned representation maintains sufficient capacity to discriminate between semantically distinct inputs. This energy-based perspective aligns with the broader JEPA framework~\cite{lecun2022path}, which advocates for learning predictive world models through energy minimization rather than explicit generative modeling or contrastive discrimination.

\subsection{Implementation Details}

As anticipated before, we use ClinicalBERT \cite{huang2020clinicalbertmodelingclinicalnotes} as our frozen text encoder, leveraging its domain-specific pretraining on clinical notes to provide semantically rich text embeddings for medical image-text pairs. A learnable projection head maps the output embeddings to the shared latent space.

As vision encoders, instead, we employ standard ViT-Base and ViT-Small architectures \cite{dosovitskiy2021imageworth16x16words}, randomly initialized with learnable weights. This choice enables direct comparison with CLIP-based finetuning baselines under matched model capacity.

\begin{algorithm}[t]
\caption{NOVA Training Step}
\label{alg:vllejepa}
\small
\texttt{vision\_enc}: ViT backbone; \texttt{pred}: 3-layer MLP; \texttt{text\_enc}: frozen ClinicalBERT + learnable projection.

\vspace{0.3em}
\begin{lstlisting}[language=Python, basicstyle=\ttfamily\scriptsize, commentstyle=\color{gray}]
def train_step(imgs, texts, lmbd=0.02):
    # multi-crop augmentation
    g_crops = global_aug(imgs)   # (B,2,C,224,224)
    l_crops = local_aug(imgs)    # (B,6,C,96,96)
    
    # encode and predict views
    g_emb = pred(vision_enc(g_crops))
    l_emb = pred(vision_enc(l_crops))
    views = cat([g_emb, l_emb], dim=0)  # (8,B,D)
    
    # text anchor (frozen backbone)
    t_emb = text_enc(texts).unsqueeze(dim=0) # (1,B,D)
    
    # alignment + regularization
    L_mse = (views - t_emb).square().mean()
    L_sig = SIGReg(cat([views, t_emb], dim=0))
    
    return (1 - lmbd) * L_mse + lmbd * L_sig
\end{lstlisting}
\end{algorithm}

\section{Experiments}

\subsection{Datasets}

\paragraph{MIMIC-CXR}
We train on MIMIC-CXR \cite{johnson2019mimic}, a large publicly available dataset of chest radiographs with free-text radiology reports collected from the Beth Israel Deaconess Medical Center (BIDMC) in Boston, MA. The dataset contains 377,110 chest X-ray images corresponding to 227,835 imaging studies from 65,379 patients presenting to the emergency department between 2011 and 2016. Each study typically includes frontal (anteroposterior or posteroanterior) and lateral views, accompanied by a semi-structured radiology report written by a practicing radiologist during routine clinical care. We filter for anteroposterior (AP) and posteroanterior (PA) views to match the predominant acquisition orientations in downstream evaluation datasets. Each image is resized to $224 \times 224$ pixels. For free-text samples, following MedCLIP \cite{wang2022medclipcontrastivelearningunpaired}, we use only the ``Impression'' section of the radiology report, which contains a concise summary of the radiologist's findings. After filtering, our training set comprises 130,291 vision-language pairs.

We use the held-out test split of MIMIC-CXR for in-distribution evaluation. This allows us to assess model performance on data from the same institution and patient population as the training set, providing a baseline for comparison with out-of-distribution generalization.

\paragraph{ChestX-ray14}
ChestX-ray14 \cite{Wang_2017}, released by the National Institutes of Health (NIH) Clinical Center, contains 112,120 frontal-view chest radiographs from 30,805 unique patients. The dataset was collected from the NIH Clinical Center and provides labels for 14 common thoracic pathologies, including Atelectasis, Cardiomegaly, Effusion, Infiltration, Mass, Nodule, Pneumonia, Pneumothorax, Consolidation, Edema, Emphysema, Fibrosis, Pleural Thickening, and Hernia. Labels were extracted from associated radiology reports using natural language processing, with an estimated accuracy exceeding 90\%. As ChestX-ray14 originates from a different institution (NIH Clinical Center vs. BIDMC) and was collected over a different time period, it serves as an out-of-distribution test set to evaluate cross-institutional generalization.

\paragraph{CheXpert}
CheXpert \cite{irvin2019chexpert}, developed by the Stanford Machine Learning Group, consists of 224,316 chest radiographs from 65,240 patients who underwent radiographic examination at Stanford Health Care between October 2002 and July 2017. The dataset includes both frontal and lateral views from inpatient and outpatient settings. A key feature of CheXpert is its handling of diagnostic uncertainty: labels are assigned as positive, negative, or uncertain using an automated rule-based labeler applied to radiology reports. The dataset defines 14 observations based on clinical relevance and report prevalence, conforming to the Fleischner Society's recommended glossary. For simplicity we treat uncertain as negative cases. As CheXpert originates from Stanford Hospital (a different institution and geographic region than BIDMC), it provides a second out-of-distribution test set to assess generalization.

\subsection{Evaluation Protocol}

We evaluate zero-shot classification performance on five pathologies from the CheXpert competition: Atelectasis, Cardiomegaly, Edema, Pleural Effusion, and Consolidation. Following MedCLIP \cite{wang2022medclipcontrastivelearningunpaired}, we frame zero-shot classification as a similarity matching problem between image and text embeddings.

\paragraph{Prompt Design} For each pathology, we construct a positive prompt (e.g., ``atelectasis'') and a negative prompt (e.g., ``no atelectasis''). Both prompts are encoded through the text encoder and projection head to obtain $\ell_2$-normalized embeddings $\mathbf{t}^+_c$ and $\mathbf{t}^-_c$ for each class $c$.

\paragraph{Zero-Shot Inference} Given a test image, we encode it through the vision encoder and predictor to obtain an $\ell_2$-normalized embedding $\mathbf{v}$. We compute cosine similarities with both positive and negative prompt embeddings:
\begin{equation}
    s^+_c = \mathbf{v}^\top \mathbf{t}^+_c, \quad s^-_c = \mathbf{v}^\top \mathbf{t}^-_c
\end{equation}
The probability of the pathology being present is computed via softmax over the positive-negative pair:
\begin{equation}
    P(y_c = 1 \mid \mathbf{v}) = \frac{\exp(s^+_c)}{\exp(s^+_c) + \exp(s^-_c)}
\end{equation}
This formulation treats each pathology as an independent binary classification problem, naturally handling the multi-label nature of chest X-ray diagnosis where multiple conditions may co-occur.

\paragraph{Metrics} We report area under the receiver operating characteristic curve (AUC) for each pathology, as well as the macro-averaged AUC across all five conditions. We report mean and standard deviation over three random seeds to assess training stability.

\subsection{Training}

We train each NOVA model for 100 epochs using AdamW with a cosine learning rate schedule, decaying from $1 \times 10^{-4}$ to $1 \times 10^{-5}$ with a single warm-up epoch. We use a batch size of 256 and generate $n=8$ augmented views per image via random augmentations. The vision encoder is initialized randomly and followed by a 3-layer MLP predictor with batch normalization and hidden dimension of $2048$. The predictor and text encoder projection head map to a shared embedding space of dimension 64. The loss weighting hyperparameter is set to $\lambda = 0.02$. We apply gradient clipping with maximum norm 1.0 and use mixed-precision training (bfloat16). We observe stable training across a wide range of hyperparameter choices. A full training run takes approximately 6 hours on a single NVIDIA H200.

\begin{table*}[t]
\centering
\small
\setlength{\tabcolsep}{6pt}
\begin{tabular}{lll cc | c | ccc}
\toprule
& & & & & & \multicolumn{3}{c}{\textbf{Zero-Shot Classification (AUC)}} \\
\cmidrule(lr){7-9}
\textbf{Framework} & \textbf{Algorithm} & \textbf{Model} & \rotatebox{90}{\textbf{\# Parameters}} & \rotatebox{90}{\textbf{\# Samples Seen}} & \rotatebox{90}{\textbf{Average}} & \textbf{MIMIC-CXR} & \textbf{Chest-Xray14} & \textbf{CheXpert} \\
\midrule
\multirow{3}{*}{CLIP} 
    & Base    & ViT-B & 150.0M & 1.28M & 46.56 & 48.61 $\pm$ 0.00 & 50.63 $\pm$ 0.00 & 40.44 $\pm$ 0.00 \\
    & InfoNCE & ViT-B & 150.0M & 1.41M & 66.29 & 66.78 $\pm$ 1.79 & 65.53 $\pm$ 1.46 & 66.56 $\pm$ 3.98 \\
    & SigLIP  & ViT-B & 150.0M & 1.41M & 68.19 & 68.49 $\pm$ 1.97 & 66.39 $\pm$ 1.47 & 69.70 $\pm$ 3.20 \\
\midrule
\multirow{2}{*}{MedCLIP} 
    &       & ViT-S & 21.7M & 130K  & 72.44 & 72.07 $\pm$ 1.10 & 67.95 $\pm$ 0.52 & 77.30 $\pm$ 0.32 \\
    &       & ViT-B & 86.0M & 130K  & 71.07 & 71.61 $\pm$ 1.11 & 66.70 $\pm$ 1.38 & 74.91 $\pm$ 1.38 \\
\midrule
\midrule
\rowcolor{tableblue}
 &  & ViT-S & 27.1M & 130K & 76.23 & 75.49 $\pm$ 0.23 & 73.04 $\pm$ 0.29 & \textbf{80.15 $\pm$ 0.08} \\
\rowcolor{tableblue}
\multirow{-2}{*}{\textbf{NOVA}} &  & ViT-B & 92.1M & 130K & \textbf{76.25} & \textbf{75.78 $\pm$ 0.15} & \textbf{73.17 $\pm$ 0.48} & 79.79 $\pm$ 0.32 \\
\bottomrule
\end{tabular}
\caption{Zero-shot performance measured by AUC ($\times 100$) across various in- and out-of-distribution datasets. Best performance is \textbf{highlighted}. Samples seen equals unique vision-language samples in the training dataset.}
\label{tab:main_results}
\end{table*}

\paragraph{Augmentation Strategy}
Following the multi-crop strategy introduced in DINO \cite{caron2021dino}, we generate multiple views per image at two resolutions to encourage local-to-global correspondence learning. For each image, we produce 2 global crops at $224 \times 224$ resolution (scale range 0.8--1.0) and 6 local crops at $96 \times 96$ resolution (scale range 0.5--0.7), yielding 8 views in total. Global crops capture near-complete views of the chest X-ray, while local crops focus on smaller anatomical regions. All views are aligned to the same text embedding, encouraging the model to learn consistent representations across scales. A local crop of a lung region should map to the same semantic space as the full radiograph described by the clinical text. Both crop types undergo identical randomized photometric augmentations: color jitter (brightness, contrast, saturation, and hue varied by $\pm 0.15$) and rotation ($\pm 10°$). Images are converted to 3-channel grayscale and normalized to zero mean and unit variance. The local crops are processed at lower resolution, reducing computational cost while increasing the effective number of views per batch.

\subsection{Benchmarks}

\paragraph{CLIP} We compare against OpenAI's pretrained CLIP ViT-Base \cite{radford2021clip} in three configurations: (1) Base, the pretrained model evaluated directly without finetuning; (2) InfoNCE, finetuned on MIMIC-CXR using the standard contrastive loss \cite{oord2019infonce}; and (3) SigLIP, finetuned on MIMIC-CXR using the sigmoid loss \cite{zhai2023siglip}. For finetuning, we use the same batch size as NOVA to ensure comparability. However, we found that finetuning a large pretrained model on our relatively small dataset required careful hyperparameter selection: we use a learning rate of $1 \times 10^{-7}$ without scheduling and a single warmup epoch, as higher learning rates led to training instability. We finetune for 10 epochs, as longer training degraded performance, likely due to overfitting to the limited training data. These difficulties highlight a practical advantage of training from scratch with NOVA, which exhibits stable optimization across a wide range of hyperparameters. 

\paragraph{MedCLIP} We additionally compare against the MedCLIP training approach \cite{wang2022medclipcontrastivelearningunpaired}, which uses InfoNCE loss with a decoupled contrastive objective. For fair comparison, we train MedCLIP-style models from scratch using the frozen ClinicalBERT text encoder and ViT-Small/ViT-Base vision encoders on the same MIMIC-CXR training set as NOVA. We use their proposed learning rate ($5 \times 10^{-5}$) and number of epochs ($10$). Longer training or higher learning rates lead to instabilities.
\\\\
Unlike NOVA's multi-crop strategy, InfoNCE and SigLIP are designed for single image-text pairs, so we apply standard augmentations to individual views aligned with their corresponding text embeddings.

\subsection{Results}

\begin{figure}[h]
  \centering
  \includegraphics[width=0.99\linewidth]{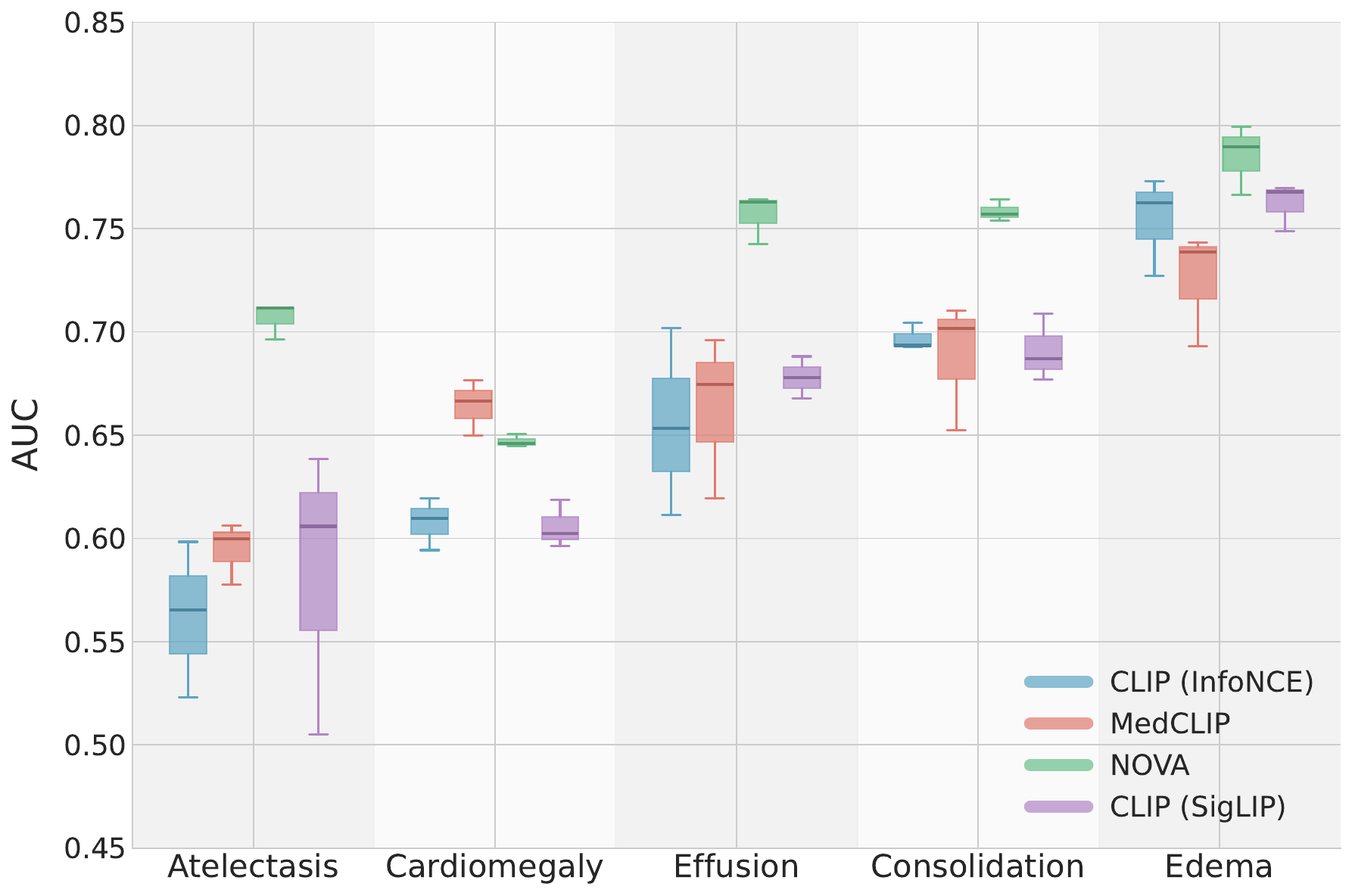}
   \caption{\textbf{ChestX-ray14 per-pathology results.} NOVA outperforms baselines across multiple conditions, with particularly strong gains on Atelectasis and Effusion.}
   \label{fig:chestxray14_results}
   \vspace{-1em}
\end{figure}

Table~\ref{tab:main_results} summarizes zero-shot classification performance across three chest X-ray benchmarks. NOVA achieves the highest average AUC of 76.25 with ViT-Base, outperforming all baselines across both in-distribution (MIMIC-CXR) and out-of-distribution (ChestX-ray14, CheXpert) evaluation sets. Under matched training conditions (same dataset, same text encoder, same vision architectures) NOVA outperforms the MedCLIP training approach by 3.8 AUC points on average (76.25 vs 72.44). The gains are particularly pronounced on out-of-distribution benchmarks, with improvements of +5.2 AUC on ChestX-ray14 and +2.9 AUC on CheXpert, suggesting that the non-contrastive objective learns representations that generalize better beyond the training distribution. The pretrained CLIP model, despite being trained on 400 million image-text pairs, achieves only 46.56 average AUC when applied directly to chest X-rays, barely above chance for some conditions. Even after finetuning on MIMIC-CXR, CLIP with SigLIP loss (68.19) underperforms NOVA by 8 AUC points, underscoring the domain gap between natural and medical images.

\begin{figure}[h]
  \centering
  \includegraphics[width=0.99\linewidth]{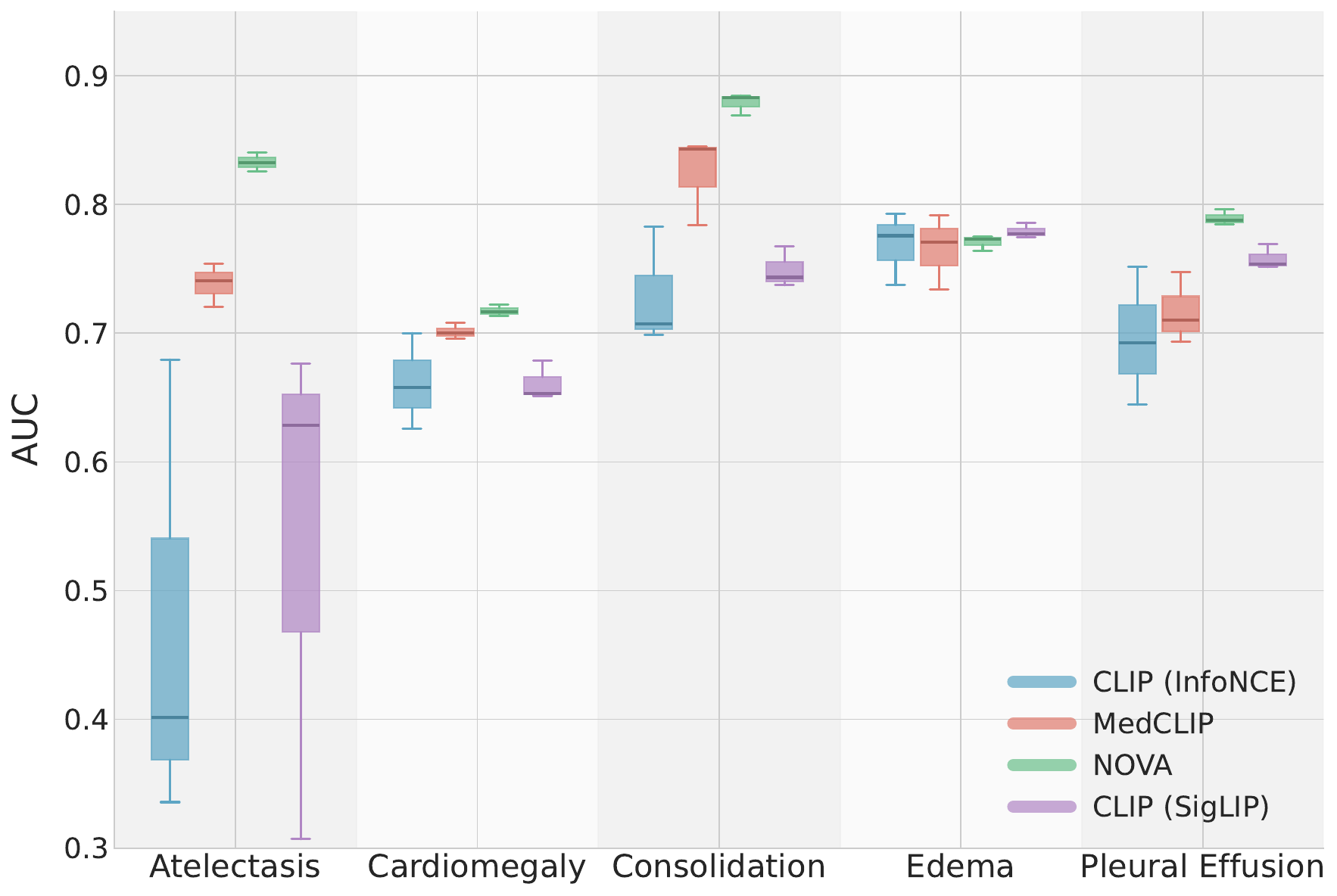}
   \caption{\textbf{CheXpert per-pathology results.} NOVA achieves strong performance on Atelectasis and Consolidation, leveraging its fine-grained vision capabilities.}
   \label{fig:chexpert_results}
   \vspace{-1em}
\end{figure}

Figures~\ref{fig:chestxray14_results} and~\ref{fig:chexpert_results} show per-pathology performance on ChestX-ray14 and CheXpert. On ChestX-ray14, NOVA demonstrates consistent improvements across all five pathologies, with the largest gains on Atelectasis (+11 AUC over MedCLIP) and Effusion (+9 AUC). On CheXpert, NOVA achieves 0.83 AUC on Atelectasis and 0.88 AUC on Consolidation, outperforming all baselines by substantial margins. These pathologies are characterized by subtle or diffuse radiographic findings that benefit from the fine-grained local-to-global alignment encouraged by our multi-crop training strategy.

\begin{figure}[t]
  \centering
  \includegraphics[width=0.99\linewidth]{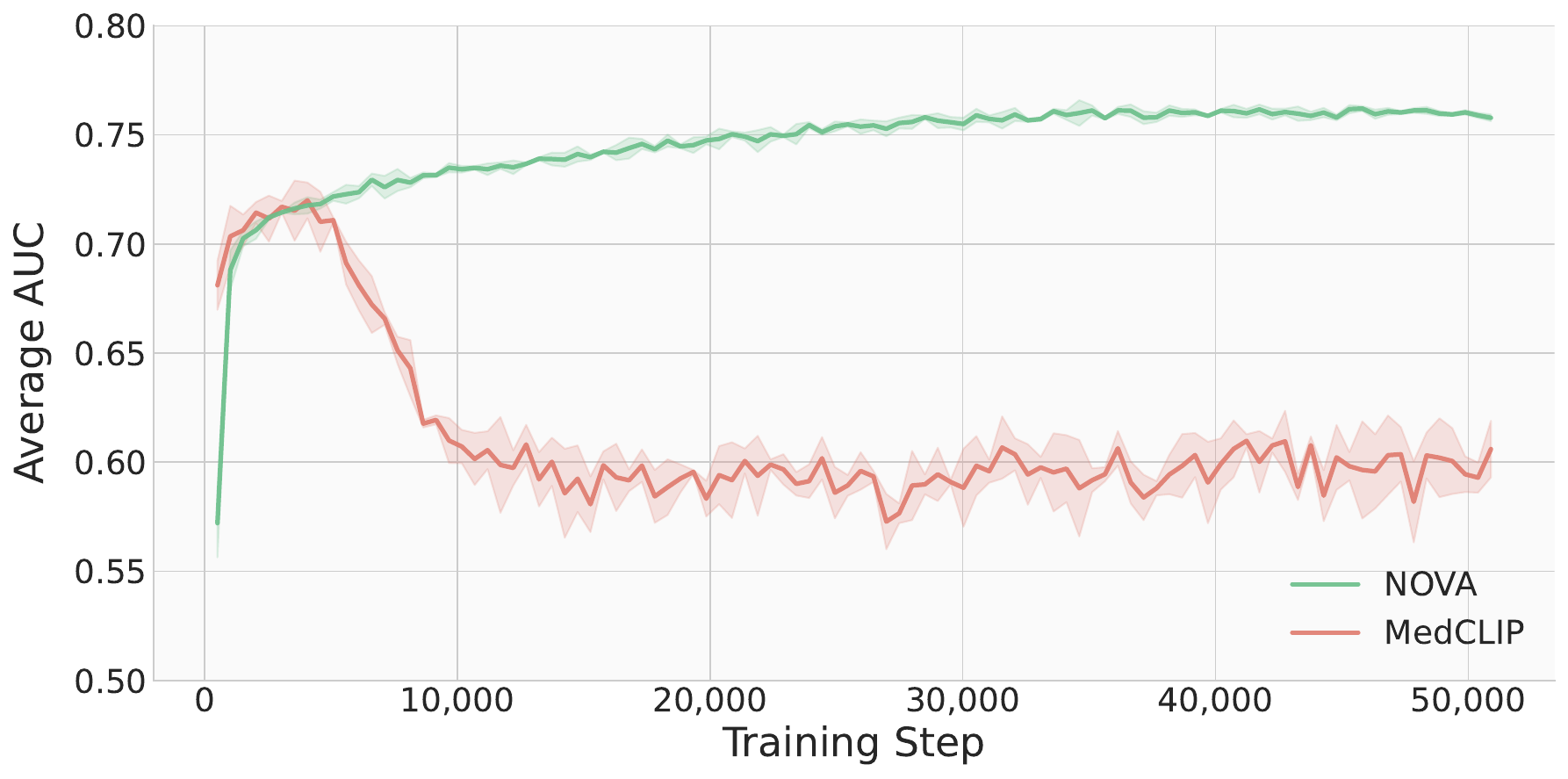}
   \caption{\textbf{Training dynamics.} NOVA converges smoothly across seeds, while MedCLIP-style training overfits after approximately 10 epochs.}
   \label{fig:zeroshot_auc_training}
   \vspace{-1em}
\end{figure}

Figure~\ref{fig:zeroshot_auc_training} shows validation AUC throughout training. NOVA exhibits smooth, monotonic convergence with tightly clustered final performance across seeds (standard deviation below $\pm$0.50 AUC). In contrast, MedCLIP-style training peaks early and degrades after approximately 10 epochs. CLIP finetuning shows even greater instability with standard deviations up to $\pm$3.98 AUC. This stability stems from SigREG's distributional regularization, which enforces consistent embedding geometry regardless of initialization and batch composition, translating to reliable results without extensive hyperparameter tuning.

NOVA also demonstrates strong parameter efficiency. The ViT-Small variant (27.1M parameters) achieves 76.23 average AUC, nearly matching ViT-Base (76.25) with 3.4$\times$ fewer parameters, and substantially outperforming the 150M parameter CLIP model. These results indicate that in low-data regimes, the inductive bias provided by the LeJEPA objective and a domain-specific text encoder is more valuable than model scale or pretraining on massive out-of-domain corpora.
\section{Conclusion}

We introduced NOVA, a non-contrastive framework for vision-language alignment that achieves state-of-the-art zero-shot classification performance on chest X-ray benchmarks while offering a simpler and more stable training objective than contrastive alternatives. On three evaluation datasets, NOVA outperforms both CLIP finetuning and MedCLIP-style training by substantial margins, with particularly strong generalization to out-of-distribution data from different institutions.

We attribute NOVA's strong performance to two factors. First, the combination of MSE alignment and SigREG regularization provides a well-conditioned optimization landscape. Unlike contrastive losses that rely on discrete positive/negative distinctions and are sensitive to batch composition, NOVA's objective is smooth and continuous, enabling stable gradient flow throughout training. Second, anchoring visual representations to ClinicalBERT, pretrained on millions of clinical notes, provides semantically meaningful targets that already capture rich medical knowledge. NOVA leverages this structure rather than learning it from scratch, allowing the vision encoder to focus on grounding visual features to established clinical concepts.

A notable finding is NOVA's robustness to hyperparameter choices. While MedCLIP-style training requires careful tuning and degrades significantly after approximately 10 epochs due to overfitting, NOVA trains stably for 100 epochs with consistent improvements. Similarly, CLIP finetuning demanded learning rates as low as $1 \times 10^{-7}$ to avoid instability, whereas NOVA tolerates learning rates two orders of magnitude higher. This robustness has practical implications: practitioners can train models without extensive hyperparameter search, reducing development time and computational cost. NOVA also trains efficiently, approximately 6 hours on a single NVIDIA H200, with the ViT-Small variant (27.1M parameters) nearly matching ViT-Base performance, suggesting suitability for resource-constrained healthcare settings.

Our evaluation is limited to chest X-ray classification across three benchmarks. While the consistent improvements across both in-distribution and out-of-distribution datasets are encouraging, further validation on other imaging modalities (CT, MRI, pathology) and tasks (segmentation, report generation) is needed to assess generalizability. A key architectural limitation is our reliance on a frozen text encoder. While this simplifies training and leverages ClinicalBERT's pretrained semantics, it may constrain the model's ability to learn vision-language alignments that deviate from the original text embedding space. An intriguing direction would be to train a domain-specific text encoder from scratch using a LeJEPA-style objective, potentially learning text representations optimized for visual grounding rather than clinical note modeling. This would require careful design to prevent collapse in the text modality, but could unlock stronger alignment for medical vision-language tasks.

Several additional directions merit exploration: extending NOVA to generative tasks such as report generation by incorporating a decoder following the VL-JEPA architecture \cite{chen2025vljepajointembeddingpredictive}, tailoring the multi-crop strategy to medical imaging through anatomically-informed cropping and investigating whether domain-specific distributional targets could further improve upon the isotropic Gaussian prior.

\section*{Acknowledgments}
This work was funded by the European Union (ERC, TAIPO, 101088594 to F.B.) grant. Views and opinions expressed are those of the authors only and do not necessarily reflect those of the European Union or ERC. Neither the European Union nor the granting authority can be held responsible for them.
{
    \small
    \bibliographystyle{ieeenat_fullname}
    \bibliography{main}

@String(CVPR= {IEEE Conf. Comput. Vis. Pattern Recog.})

@String(AAAI = {AAAI})

@String(CVPR  = {CVPR})

@misc{radford2021clip,
      title={Learning Transferable Visual Models From Natural Language Supervision}, 
      author={Alec Radford and Jong Wook Kim and Chris Hallacy and Aditya Ramesh and Gabriel Goh and Sandhini Agarwal and Girish Sastry and Amanda Askell and Pamela Mishkin and Jack Clark and Gretchen Krueger and Ilya Sutskever},
      year={2021},
      eprint={2103.00020},
      archivePrefix={arXiv},
      primaryClass={cs.CV},
      url={https://arxiv.org/abs/2103.00020}, 
}

@misc{wang2022medclipcontrastivelearningunpaired,
      title={MedCLIP: Contrastive Learning from Unpaired Medical Images and Text}, 
      author={Zifeng Wang and Zhenbang Wu and Dinesh Agarwal and Jimeng Sun},
      year={2022},
      eprint={2210.10163},
      archivePrefix={arXiv},
      primaryClass={cs.CV},
      url={https://arxiv.org/abs/2210.10163}, 
}

@misc{oord2019infonce,
      title={Representation Learning with Contrastive Predictive Coding}, 
      author={Aaron van den Oord and Yazhe Li and Oriol Vinyals},
      year={2019},
      eprint={1807.03748},
      archivePrefix={arXiv},
      primaryClass={cs.LG},
      url={https://arxiv.org/abs/1807.03748}, 
}

@misc{zhai2023siglip,
      title={Sigmoid Loss for Language Image Pre-Training}, 
      author={Xiaohua Zhai and Basil Mustafa and Alexander Kolesnikov and Lucas Beyer},
      year={2023},
      eprint={2303.15343},
      archivePrefix={arXiv},
      primaryClass={cs.CV},
      url={https://arxiv.org/abs/2303.15343}, 
}

@misc{balestriero2025lejepa,
      title={LeJEPA: Provable and Scalable Self-Supervised Learning Without the Heuristics}, 
      author={Randall Balestriero and Yann LeCun},
      year={2025},
      eprint={2511.08544},
      archivePrefix={arXiv},
      primaryClass={cs.LG},
      url={https://arxiv.org/abs/2511.08544}, 
}

@misc{chen2025vljepajointembeddingpredictive,
      title={VL-JEPA: Joint Embedding Predictive Architecture for Vision-language}, 
      author={Delong Chen and Mustafa Shukor and Theo Moutakanni and Willy Chung and Jade Yu and Tejaswi Kasarla and Allen Bolourchi and Yann LeCun and Pascale Fung},
      year={2025},
      eprint={2512.10942},
      archivePrefix={arXiv},
      primaryClass={cs.CV},
      url={https://arxiv.org/abs/2512.10942}, 
}

@misc{dosovitskiy2021imageworth16x16words,
      title={An Image is Worth 16x16 Words: Transformers for Image Recognition at Scale}, 
      author={Alexey Dosovitskiy and Lucas Beyer and Alexander Kolesnikov and Dirk Weissenborn and Xiaohua Zhai and Thomas Unterthiner and Mostafa Dehghani and Matthias Minderer and Georg Heigold and Sylvain Gelly and Jakob Uszkoreit and Neil Houlsby},
      year={2021},
      eprint={2010.11929},
      archivePrefix={arXiv},
      primaryClass={cs.CV},
      url={https://arxiv.org/abs/2010.11929}, 
}

@misc{huang2020clinicalbertmodelingclinicalnotes,
      title={ClinicalBERT: Modeling Clinical Notes and Predicting Hospital Readmission}, 
      author={Kexin Huang and Jaan Altosaar and Rajesh Ranganath},
      year={2020},
      eprint={1904.05342},
      archivePrefix={arXiv},
      primaryClass={cs.CL},
      url={https://arxiv.org/abs/1904.05342}, 
}

@article{bardes2021vicreg,
  title={Vicreg: Variance-invariance-covariance regularization for self-supervised learning},
  author={Bardes, Adrien and Ponce, Jean and LeCun, Yann},
  journal={arXiv preprint arXiv:2105.04906},
  year={2021}
}

@inproceedings{eslami2023pubmedclip,
  title={Pubmedclip: How much does clip benefit visual question answering in the medical domain?},
  author={Eslami, Sedigheh and Meinel, Christoph and De Melo, Gerard},
  booktitle={Findings of the Association for Computational Linguistics: EACL 2023},
  pages={1181--1193},
  year={2023}
}

@article{zhang2023biomedclip,
  title={Biomedclip: a multimodal biomedical foundation model pretrained from fifteen million scientific image-text pairs},
  author={Zhang, Sheng and Xu, Yanbo and Usuyama, Naoto and Xu, Hanwen and Bagga, Jaspreet and Tinn, Robert and Preston, Sam and Rao, Rajesh and Wei, Mu and Valluri, Naveen and others},
  journal={arXiv preprint arXiv:2303.00915},
  year={2023}
}

@article{lecun2022path,
  title={A path towards autonomous machine intelligence version 0.9. 2, 2022-06-27},
  author={LeCun, Yann},
  journal={Open Review},
  volume={62},
  number={1},
  pages={1--62},
  year={2022}
}

@article{Shurrab_2022,
   title={Self-supervised learning methods and applications in medical imaging analysis: a survey},
   volume={8},
   ISSN={2376-5992},
   url={http://dx.doi.org/10.7717/peerj-cs.1045},
   DOI={10.7717/peerj-cs.1045},
   journal={PeerJ Computer Science},
   publisher={PeerJ},
   author={Shurrab, Saeed and Duwairi, Rehab},
   year={2022},
   month=jul, pages={e1045} }

@article{huang2023self,
  title={Self-supervised learning for medical image classification: a systematic review and implementation guidelines},
  author={Huang, Shih-Cheng and Pareek, Anuj and Jensen, Malte and Lungren, Matthew P and Yeung, Serena and Chaudhari, Akshay S},
  journal={NPJ Digital Medicine},
  volume={6},
  number={1},
  pages={74},
  year={2023},
  publisher={Nature Publishing Group UK London}
}

@inproceedings{chen2020simple,
  title={A simple framework for contrastive learning of visual representations},
  author={Chen, Ting and Kornblith, Simon and Norouzi, Mohammad and Hinton, Geoffrey},
  booktitle={International conference on machine learning},
  pages={1597--1607},
  year={2020},
  organization={PmLR}
}

@article{grill2020bootstrap,
  title={Bootstrap your own latent-a new approach to self-supervised learning},
  author={Grill, Jean-Bastien and Strub, Florian and Altch{\'e}, Florent and Tallec, Corentin and Richemond, Pierre and Buchatskaya, Elena and Doersch, Carl and Avila Pires, Bernardo and Guo, Zhaohan and Gheshlaghi Azar, Mohammad and others},
  journal={Advances in neural information processing systems},
  volume={33},
  pages={21271--21284},
  year={2020}
}

@inproceedings{he2020momentum,
  title={Momentum contrast for unsupervised visual representation learning},
  author={He, Kaiming and Fan, Haoqi and Wu, Yuxin and Xie, Saining and Girshick, Ross},
  booktitle={Proceedings of the IEEE/CVF conference on computer vision and pattern recognition},
  pages={9729--9738},
  year={2020}
}

@misc{sowrirajan2021mococxrmocopretrainingimproves,
      title={MoCo-CXR: MoCo Pretraining Improves Representation and Transferability of Chest X-ray Models}, 
      author={Hari Sowrirajan and Jingbo Yang and Andrew Y. Ng and Pranav Rajpurkar},
      year={2021},
      eprint={2010.05352},
      archivePrefix={arXiv},
      primaryClass={cs.CV},
      url={https://arxiv.org/abs/2010.05352}, 
}

@misc{azizi2021bigselfsupervisedmodelsadvance,
      title={Big Self-Supervised Models Advance Medical Image Classification}, 
      author={Shekoofeh Azizi and Basil Mustafa and Fiona Ryan and Zachary Beaver and Jan Freyberg and Jonathan Deaton and Aaron Loh and Alan Karthikesalingam and Simon Kornblith and Ting Chen and Vivek Natarajan and Mohammad Norouzi},
      year={2021},
      eprint={2101.05224},
      archivePrefix={arXiv},
      primaryClass={eess.IV},
      url={https://arxiv.org/abs/2101.05224}, 
}

@article{johnson2019mimic,
  title={MIMIC-CXR, a de-identified publicly available database of chest radiographs with free-text reports},
  author={Johnson, Alistair EW and Pollard, Tom J and Berkowitz, Seth J and Greenbaum, Nathaniel R and Lungren, Matthew P and Deng, Chih-ying and Mark, Roger G and Horng, Steven},
  journal={Scientific data},
  volume={6},
  number={1},
  pages={317},
  year={2019},
  publisher={Nature Publishing Group UK London}
}

@inproceedings{Wang_2017,
   title={ChestX-Ray8: Hospital-Scale Chest X-Ray Database and Benchmarks on Weakly-Supervised Classification and Localization of Common Thorax Diseases},
   url={http://dx.doi.org/10.1109/CVPR.2017.369},
   DOI={10.1109/cvpr.2017.369},
   booktitle={2017 IEEE Conference on Computer Vision and Pattern Recognition (CVPR)},
   publisher={IEEE},
   author={Wang, Xiaosong and Peng, Yifan and Lu, Le and Lu, Zhiyong and Bagheri, Mohammadhadi and Summers, Ronald M.},
   year={2017},
   month=jul, pages={3462–3471} }

@inproceedings{irvin2019chexpert,
  title={Chexpert: A large chest radiograph dataset with uncertainty labels and expert comparison},
  author={Irvin, Jeremy and Rajpurkar, Pranav and Ko, Michael and Yu, Yifan and Ciurea-Ilcus, Silviana and Chute, Chris and Marklund, Henrik and Haghgoo, Behzad and Ball, Robyn and Shpanskaya, Katie and others},
  booktitle={Proceedings of the AAAI conference on artificial intelligence},
  volume={33},
  number={01},
  pages={590--597},
  year={2019}
}

@misc{devlin2019bertpretrainingdeepbidirectional,
      title={BERT: Pre-training of Deep Bidirectional Transformers for Language Understanding}, 
      author={Jacob Devlin and Ming-Wei Chang and Kenton Lee and Kristina Toutanova},
      year={2019},
      eprint={1810.04805},
      archivePrefix={arXiv},
      primaryClass={cs.CL},
      url={https://arxiv.org/abs/1810.04805}, 
}

@article{lee2019biobert,
   title={BioBERT: a pre-trained biomedical language representation model for biomedical text mining},
   volume={36},
   ISSN={1367-4811},
   url={http://dx.doi.org/10.1093/bioinformatics/btz682},
   DOI={10.1093/bioinformatics/btz682},
   number={4},
   journal={Bioinformatics},
   publisher={Oxford University Press (OUP)},
   author={Lee, Jinhyuk and Yoon, Wonjin and Kim, Sungdong and Kim, Donghyeon and Kim, Sunkyu and So, Chan Ho and Kang, Jaewoo},
   editor={Wren, Jonathan},
   year={2019},
   month=sep, pages={1234–1240} }

@article{gu2021pubmedpert,
   title={Domain-Specific Language Model Pretraining for Biomedical Natural Language Processing},
   volume={3},
   ISSN={2637-8051},
   url={http://dx.doi.org/10.1145/3458754},
   DOI={10.1145/3458754},
   number={1},
   journal={ACM Transactions on Computing for Healthcare},
   publisher={Association for Computing Machinery (ACM)},
   author={Gu, Yu and Tinn, Robert and Cheng, Hao and Lucas, Michael and Usuyama, Naoto and Liu, Xiaodong and Naumann, Tristan and Gao, Jianfeng and Poon, Hoifung},
   year={2021},
   month=oct, pages={1–23} }

@misc{caron2021dino,
      title={Emerging Properties in Self-Supervised Vision Transformers}, 
      author={Mathilde Caron and Hugo Touvron and Ishan Misra and Hervé Jégou and Julien Mairal and Piotr Bojanowski and Armand Joulin},
      year={2021},
      eprint={2104.14294},
      archivePrefix={arXiv},
      primaryClass={cs.CV},
      url={https://arxiv.org/abs/2104.14294}, 
}

@article{lecun2006tutorial,
  title={A tutorial on energy-based learning},
  author={LeCun, Yann and Chopra, Sumit and Hadsell, Raia and Ranzato, M and Huang, Fujie and others},
  journal={Predicting structured data},
  volume={1},
  number={0},
  year={2006}
}
}


\end{document}